\documentclass[10pt,journal,letterpaper]{IEEEtran}

\usepackage{amsmath,cite,psfrag,subfigure,graphicx}
\usepackage{algorithm}
\usepackage{algorithmic}
\begin{document}
\graphicspath{{figures/}}
\newcommand{\sps}{\scriptsize}
\newcommand{\eqs}{\footnotesize}
\title{Dirichlet Mixture Model based VQ Performance Prediction for Line Spectral Frequency}

\author{Zhanyu~Ma
\IEEEcompsocitemizethanks{\IEEEcompsocthanksitem Zhanyu Ma is with the Pattern Recognition and Intelligent System Laboratory,
Beijing University of Posts and Telecommunications, Beijing, China.\protect\\}}

\IEEEcompsoctitleabstractindextext{
\begin{abstract}
In this paper, we continue our previous work on the Dirichlet mixture model (DMM)-based VQ to derive the performance bound of the LSF VQ. The LSF parameters are transformed into the $\Delta$LSF domain and the underlying distribution of the $\Delta$LSF parameters are modelled by a DMM with finite number of mixture components. The quantization distortion, in terms of the mean squared error (MSE), is calculated with the high rate theory. The mapping relation between the perceptually motivated log spectral distortion (LSD) and the MSE is empirically approximated by a polynomial. With this mapping function, the minimum required bit rate for transparent coding of the LSF is estimated.
\end{abstract}

}

\maketitle

\IEEEdisplaynotcompsoctitleabstractindextext

\IEEEpeerreviewmaketitle

\section{Introduction}
\IEEEPARstart{F}{or} the purpose of efficient speech transmission, the linear predictive coding (LPC) coefficients are usually converted to other representation forms~\cite{Paliwal1995,Ma2010e,Vary2006,Ma2012}. The line spectral frequency (LSF) representation, among others, is the commonly used representation in the LPC model transmission~\cite{Kleijn2003,Paliwal1995, Itakura1975,
Paliwal1993, 09,Soong1993,Ma2010d} because it has a relatively uniform spectral sensitivity~\cite{Gardner1995,Li1999,Ma2010a}. From the speech quality point of view, the log spectral distortion (LSD) is the most preferred objective distortion measure in the literature~\cite{Kleijn2007,Ekman2008,Chatterjee2008}. It has been recognized that~\cite{Paliwal1993,Ma2010}, with $1$dB average spectral distortion and constraints on the outliers, the transparent quality of the reconstructed speech signal is guaranteed. However, minimizing the LSD directly is analytically intractable and computationally costly. Therefore, it is of great interest to study the method which can calculate the minimum bit rate that fulfills the transparent coding requirement.

Recently, probability density function (PDF)-optimized VQ~\cite{Hedelin2000,Ma2009,Subramaniam2003,Ma2010a,Ma2013} draws more and more attentions. The PDF-optimized VQ proposes a parametric VQ strategy where the underlying distribution of the LSF parameters are described by a mixture of distributions, such as the Gaussian mixture model (GMM)~\cite{Subramaniam2003,08}, the beta mixture model (BMM)~\cite{Ma2010a}, and the Dirichlet mixture model (DMM)~\cite{Ma2013}. One advantage of such parametric VQ method is that, with the high rate assumption~\cite{Gray1990,Ma2008,Kleijn2010,Ma2013}, the distortion-rate (D-R) relation is analytically tractable, where the distortion is defined as the mean squared error (MSE). In the high rate vector quantization (VQ) case, a VQ using an MSE as distortion criterion can approach the performance of a VQ designed using the LSD criterion~\cite{Gardner1995,Li1999,Ma2015a}. Based on this fact, Chatterjee et al.~\cite{Chatterjee2008} proposed a method to predict the LSF VQ performance bound. In~\cite{Chatterjee2008}, the distribution of the LSF parameters were modelled by a GMM. In order to describe the relation between the MSE and the LSD, a third-order polynomial was fitted to provide the mapping of MSE to LSD. The MSE and the corresponding LSD were obtained by implementing the split VQ strategy~\cite{Paliwal1993}.

The LSF parameters have a number of properties: the support range is bounded, the parameters are ordered, and the filter stability can be easily checked~\cite{Paliwal1995}. It has been shown that by exploiting such properties, the VQ performance of the LSF parameters can be significantly improved~\cite{Lindblom2003,Ma2010a,Ma2013}. A bounded-support GMM-based VQ was proposed in~\cite{Lindblom2003}, where the Gaussian distribution was truncated to fit the bounded support property of the LSF parameters. For the purpose of reducing the computational cost in the truncated GMM, we introduced another bounded-support parametric VQ, the BMM-based VQ, in~\cite{Ma2010a}. The BMM-based VQ also considered the bounded-support nature of the LSF parameters. To further exploit the ordering property, we modeled the underlying distribution of the LSF parameters by a DMM and proposed a DMM-based VQ~\cite{Ma2013}. The LSF parameters were transformed to the $\Delta$LSF representation where the bounded and ordering properties were exploited explicitly. The analytically tractable expression of the MSE via the DMM-based VQ was derived. Since the DMM-based VQ outperforms the GMM-based VQ, it potentially permits a lower LSF VQ performance bound.

Inspired by the idea in~\cite{Chatterjee2008}, we study the performance bound of the DMM-based LSF VQ in this paper. The polynomial derived in~\cite{Chatterjee2008} is still used to map the MSE to LSD, as we consider the relation between the MSE and the LSD is VQ independent. With the calculated MSE expression, the performance bound of the DMM-based LSF VQ is obtained.

The rest of this paper is organized as follows: we review the $\Delta$LSF representation and the DMM modeling in section~\ref{Chap DMM modeling}. The D-R relation of the parametric VQ is presented in section~\ref{Chap DR relation}. The performance bound estimation is carried out in section~\ref{Chap performance bound} and some conclusions are drawn in section~\ref{Chap conclusion}.

\section{$\Delta$LSF representation and modelling}
\label{Chap DMM modeling}
The representation of LPC parameters by LSF was introduced by Itakura~\cite{Itakura1975}, and LSFs
are widely used for speech coding because of the advantage compared to some other forms of representations (such as LARs, ASRCs). By
taking a linear predictive model with order $K$, the LSFs
In the linear predictive coding model, the filter $G(z)$ with
order $K$ is
\begin{equation}
\eqs
G(z) = 1 + \sum_{k=1}^K a_k z^{-k}.
\vspace{-0mm}
\end{equation}
Then we can build a symmetric polynomial
\begin{equation}
\eqs
\vspace{-0mm}
P(z) = G(z) + z^{-(K+1)}G(z^{-1})
\vspace{-0mm}
\end{equation}
and an anti-symmetric polynomial
\begin{equation}
\eqs
Q(z) = G(z) - z^{-(K+1)}G(z^{-1}).
\vspace{0mm}
\end{equation}
The zeros of $P(z)$ and $Q(z)$
are interleaved on the unit circle~\cite{Vary2006,Soong1984} as
\begin{equation}
\eqs
{0=\omega_{q_0}<\omega_{p_1}< \omega_{q_1}<\ldots<\omega_{q_{\frac{K}{2}}}<\omega_{p_{\frac{K}{2}+1} }= \pi}.
\vspace{0mm}
\end{equation}
Then the LSF parameters are obtained as
\begin{equation}
\eqs
\mathbf{s} = [s_1,s_2,\ldots,s_K]^{\text{T}} = [\omega_{p_1}, \omega_{q_1},\ldots,\omega_{p_{\frac{K}{2}}},\omega_{q_{\frac{K}{2}}}]^{\text{T}}.
\vspace{-2mm}
\end{equation}
Since the LSF parameters are in the interval $(0,\pi)$ and are
strictly ordered, we represent the LSF parameters with another
representation named $\Delta$LSF defined as~\cite{Ma2010}
\begin{equation}
\eqs
\label{dLSF}
\mathbf{x}=\frac{1}{\pi}[x_1,x_2,\ldots,x_K]^{\text{T}} = \frac{1}{\pi}[s_1,s_2-s_1,\ldots,s_K-s_{K-1}]^{\text{T}}.
\vspace{-0mm}
\end{equation}
With a transformation matrix $A$, the relation between the LSF
parameters and $\Delta$LSF parameters can be denoted as
\begin{equation}
\eqs
\label{TransformationFunc}
\textbf{x} = \varphi(\mathbf{s})=A\textbf{s},
\vspace{-3mm}
\end{equation}
where
\begin{equation}
\eqs
\nonumber
A = \frac{1}{\pi} \left[\begin{array}{cccccc}
1 & 0 &\cdots &\cdots & \cdots&0 \\
-1 & 1 & 0 & \cdots & \cdots & 0 \\
0 & -1 & 1 & 0 & \cdots & 0 \\
\vdots & \vdots & \ddots & \ddots & \ddots & \vdots \\
%0 & \cdots & 0 & -1 & 1 & 0 \\
0 & \cdots & \cdots & 0 & -1 & 1 \\
\end{array}\right]_{K\times K}.
\end{equation}
From the definition of the $\Delta$LSF, we know that $x_k>0,\ k=1,\dots,K$ and $\sum_{k=1}^{K}x_k<1$. By introducing $x_{K+1} = 1
- \sum_{k=1}^{K} x_k$, it is reasonable to model the PDF of vector $\mathbf{x} = [x_1,\ldots,x_K]^{\text{T}}$ by the Dirichlet
distribution with $K+1$ parameters. The approach of mixture models assumes that the observed
samples are drawn from a mixture of parametric distributions. With a set of $N$ $\emph{i.i.d.}$ observations
$\mathbf{X}=[\mathbf{x}_1,\ldots,\mathbf{x}_N]$, we can denote the likelihood function for the observations by a DMM with $I$ components as
\begin{equation}
\eqs
\begin{split}
f(\mathbf{X}) &= \prod_{n=1}^{N}\sum_{i=1}^{I}\pi_i \mathbf{Dir}(\mathbf{x}_n;\boldsymbol{\alpha}_i)\\&=\prod_{n=1}^{N}\sum_{i=1}^{I}\pi_i\frac{\Gamma(\sum_{k=1}^{K+1}\alpha_{ki})}{\prod_{k=1}^{K+1}\Gamma(\alpha_{ki})}\prod_{k=1}^{K+1}x_{kn}^{\alpha_{ki}-1},
\end{split}
\end{equation}
where $\Gamma(\cdot)$ is the gamma function, $\boldsymbol{\alpha}_i=[\alpha_{1i},\alpha_{2i},\ldots,\alpha_{K+1,i}]^{\text{T}}$ is the parameter vector for the $i$th mixture
component, and $\pi_i$ is the nonnegative weighting factor for the $i$th component, and $\sum_{i=1}^I{\pi_i}=1$. By applying the
expectation-maximization (EM) algorithm, the parameters in the DMM can be estimated as in~\cite{Ma2010}. %by iteratively computing:
%\begin{equation}
%\nonumber
%\label{EM for DMM}
%\begin{split}
%&\textnormal{E step:}\\
%&\overline{z}_{ni} =\mathbf{E} [z_{ni}] = \frac{\pi_i \mathbf{Dir}(\mathbf{x}_n;\boldsymbol{\alpha}_i)}{\sum_{j=1}^I \pi_{j}\mathbf{Dir}(\mathbf{x}_n;\boldsymbol{\alpha}_j)}\\
%&\textnormal{M step:}\\
%&\pi_i=\frac{1}{N}\sum_{n=1}^N\overline{z}_{ni}\\
%&\left [\begin{array}{c}
%\psi(\alpha_{1i})-\psi(\sum_{k=1}^{K+1}\alpha_{ki})\\
%\vdots\\
% \psi(\alpha_{K+1,i})-\psi(\sum_{k=1}^{K+1}\alpha_{ki})\\
%\end{array}\right]=\left [\begin{array}{c}
%\frac{\sum_{n=1}^N
%\overline{z}_{ni} \log x_{1n}}{\sum_{n=1}^N
%\overline{z}_{ni}}\\
%\vdots\\
%\frac{\sum_{n=1}^N
%\overline{z}_{ni} \log x_{K+1,n}}{\sum_{n=1}^N
%\overline{z}_{ni}}\\
%\end{array}\right].
%\end{split}
%\end{equation}
\section{Distortion rate performance via DMM}
\label{Chap DR relation}
In this section, we will review the fundamental theory of vector quantization in the high rate case. The D-R performance of the LSF VQ via DMM modeling will also be introduced. More details of the basis of high rate theory can be found in,~\emph{e.g.},~\cite{Gray1990,Kleijn2010}.
\subsection{Distortion with centroid density}
For quantization purpose, we consider the distortion measured by the weighted squared error as~\cite{Gardner1995}
\begin{equation}
\eqs
\label{WSE}
d_{\text{W}}(\mathbf{x},\mathbf{\hat{x}})= \frac{1}{2}(\mathbf{x} - \mathbf{\hat{x}})^{\text{T}}\mathbf{W}(\mathbf{x} - \mathbf{\hat{x}}),
\end{equation}
where $\mathbf{\hat{x}}$ is the reconstruction point of $\mathbf{x}$ and $\mathbf{W}$ is named as the ``sensitivity" matrix which is positive definite. Let $f_{{\mathbf{x}}}(\mathbf{x})$ denote the PDF of the $K$-dimensional $\Delta$LSF parameter vector and consider a vector quantizer with $L$ cells $\Omega_l$ centered at $\mathbf{c}_l$. With the above notation, the weighted MSE (WMSE) for each cell, on a per dimension basis, is calculated as~\cite{Kleijn2010}
\begin{equation}
\eqs
\label{WMSE}
\begin{split}
D_{l}&=\frac{1}{K}\frac{\int_{\Omega_l}f_{\mathbf{x}}(\mathbf{x})d_{\text{W}}(\mathbf{x},\mathbf{\hat{x}})d\mathbf{x}}{\int_{\Omega_l}f_{\mathbf{x}}(\mathbf{x})d\mathbf{x}}\\
&\approx \frac{1}{KV_l}\int_{\Omega_l}f_{\mathbf{x}}(\mathbf{x})d_{\text{W}}(\mathbf{x},\mathbf{c}_l)d\mathbf{x},\\
%&=V_l^{\frac{2}{K}}C(2,K,\mathcal{G}(l)),
\end{split}
\end{equation}
where $V_l$ is the volume of the cell $\Omega_l$. To measure the distortion of quantizing the LSF parameters, the WMSE was proven to be equivalent to the LSD
measurement asymptotically, with high rate~\cite{Li1999}. However, the weighted coefficient for each dimension in this quadratic distortion measure is signal dependent and, therefore, deriving an analytically tractable expression for~\eqref{WMSE} is mathematically intractable. Hence, the plain MSE, which is the WMSE with the same weights for all the dimensions, in usually applied to simplify the derivation so that an analytically tractable expression can be obtained. In such case, the ``sensitivity" matrix $\mathbf{W}$ in~\eqref{WSE} is replaced by $2\mathbf{I}$, where $\mathbf{I}$ is the identity matrix. Then the quantization distortion for each cell in~\eqref{WMSE} is approximately calculated as
\begin{equation}
\eqs
\label{MSE}
\begin{split}
D_{l}&\approx \frac{1}{KV_l}\int_{\Omega_l}f_{\mathbf{x}}(\mathbf{x})d(\mathbf{x},\mathbf{c}_l)d\mathbf{x}=V_l^{\frac{2}{K}}C(2,K,\mathcal{G}(l))
\end{split}
\end{equation}
where $d(\mathbf{x},\mathbf{c}_l)=(\mathbf{x} - \mathbf{\hat{x}})^{\text{T}}(\mathbf{x} - \mathbf{\hat{x}})$, $C(2,K,\mathcal{G}(l))$ is the coefficient of quantization defined as
\begin{equation}
C(2,K,\mathcal{G}(l))=\frac{1}{K}\frac{1}{V_l^{\frac{K+2}{K}}}\int_{\Omega_l}f_{\mathbf{x}}(\mathbf{x})d(\mathbf{x},\mathbf{c}_l)d\mathbf{x},
\end{equation}
and $\mathcal{G}(l)$ indicates the geometry of cell $l$. When the PDF is sufficiently smooth and the number of cells is sufficiently large, we can replace $\mathcal{G}(l)$ by $\mathcal{G}(\mathbf{x})$ and the function $C(2,K,\mathcal{G}(\mathbf{x}))$ is the so-called inertial profile~\cite{Na1995}. It is commonly assumed that the coefficient of quantization is not varying with the cell index in the optimal geometry case. Then we have the total quantization distortion as
\begin{equation}
\eqs
\label{TotalDistortion}
\begin{split}
D&=\sum_{l\in L} p_{L}(l)D_l\\
&\approx  \sum_{l\in L} p_L(l) V_l^{\frac{2}{K}}C(2,K,\mathcal{G})\\
&\approx C(2,K,\mathcal{G}) \int_{R^K} f_{\mathbf{x}}(\mathbf{x})g_C(\mathbf{x})^{-\frac{2}{K}}d\mathbf{x},
\end{split}
\end{equation}
where $p_L(l)$ is the probability mass of the $l$th cell and $g_C(\mathbf{x})$ is the density of the quantization centroid. In~\eqref{TotalDistortion}, the coefficient of quantization $C(2,K,\mathcal{G})$ is $\frac{1}{\pi}\frac{K}{K+2}\left(\frac{K}{2}\Gamma(\frac{K}{2})\right)^{\frac{2}{K}}$, which only depends on the dimensionality.
\begin{table}[!t]
\caption{\label{Tab:Polynomial}\small Coefficients of the third-order polynomial.}
\centering
\begin{tabular}{|c|c|c|c|c|}
\hline
Coefficient & $c_0$ & $c_1$ & $c_2$ & $c_3$\\
\hline
Value (in $\times10^5$ )& $0.0000$&$0.0023$&$-0.1291$&$3.7704$\\
\hline
\end{tabular}
\vspace{-4mm}
\end{table}
\subsection{D-R performance of DMM-based VQ}
Assuming the average rate used for quantization is $R$ bits/vector, with the high rate theory and in the constrained entropy case, we have the optimal inter-component bit allocation strategy under the DMM modeling as~\cite{Ma2010}
\begin{equation}
\eqs
R_i = R_q + h_i(\mathbf{x}) - \sum_{i=1}^I \pi_ih_i(\mathbf{x}),
\end{equation}
where $R_i$ denotes the rate assigned to the $i$th component, $R_q=R-\log_2I$ is the rate spent on quantizing the signal, $h_i(\mathbf{x})$ is the differential entropy of the $i$th mixture component. Please note, with such expression, we assume that there is no overlap among different mixture components. In the constrained entropy case, the centroid densities for all the mixture components are identical to each other~\cite{Kleijn2010}. Furthermore, the centroid density is constant so that it does not depend on the PDF of $\mathbf{x}$. According to~\eqref{TotalDistortion}, the distortions incurred by all the mixture components are also identical. Therefore, the overall distortion obtained from the DMM-based VQ is the same as the distortion incurred by any mixture component. This relation can be expressed as
\begin{equation}
\eqs
\label{DRrelation}
D_{\text{tot}}=D_i=C(2,K,\mathcal{G})2^{-\frac{2}{K}(R_q  - \sum_{i=1}^I \pi_ih_i(\mathbf{x}))},\ i=1,\ldots,I.
\end{equation}
\subsection{Mapping MSE to LSD}
The LPC model represents the speech intelligibility. From the speech quality point of view, LSD is the most preferred objective measurement of distortion~\cite{Kleijn2007}. For the $n$th frame, the LSD is defined as
\begin{equation}
\eqs
{LSD}_n=\sqrt{\frac{1}{F_s}\int_0^{F_s}\left[10\log_{10}P_n(f) - 10\log_{10}\widehat{P}_n({f})\right]^2df},
\end{equation}
where $n$ is the index of the vector, $F_s$ is the sampling frequency in Hz, $P_n(f)$ and $\widehat{P}_n({f})$ are the original and quantized LPC power spectra
of the $n$th vector. $P(f)$ and $\widehat{P}({f})$ are calculated as
\begin{equation}
\eqs
\begin{split}
P_n(f) &= 1/|A_n(e^{j2\pi f/F_s})|^2,\ A(z) = 1 + \sum_{k=1}^K a_k z^{-k}\\
\widehat{P}_n({f}) &= 1/|\widehat{A}_n(e^{j2\pi f/F_s})|^2,\ \widehat{A}(z) = 1 + \sum_{k=1}^K \widehat{a}_k z^{-k}\\
\end{split}
\end{equation}
where $a_k,\ k=1,\ldots,K$ are the corresponding LPC parameters. The evaluation criteria for the VQ is
\begin{enumerate}
\item $1$ dB LSD on average,
\item less than $2\%$ outliers in $2-4$ dB range,
\item no outlier larger than $4$ dB.
\end{enumerate}
The LSD was originally proposed for narrow-band speech. It has been demonstrated~\cite{Guibe2001,So2006,Ekman2008} that the LSD is also a suitable distortion measure for wide-band speech. In general, directly minimizing the LSD is computationally costly, as it is difficult to obtain the VQ centroid and the search complexity is high. In practice, it is feasible to get an analytically tractable expression for the VQ's D-R performance (as shown in~\eqref{DRrelation}). Thus, we map the MSE to the LSD empirically so that a closed-form expression between the LSD and the bit rate can be obtained as well. In~\cite{Chatterjee2008}, the five-split VQ method~\cite{Chatterjee2007} was applied to get the MSE and the corresponding LSD values, at different bit rate. Then a third-order polynomial was fitted to get the mapping function. Assuming that the mapping from the MSE to the LSD is method independent, we continue to use that third-order polynomial in this paper. The coefficients of the third-order polynomial are listed in Table~\ref{Tab:Polynomial}.

\section{Performance bound estimation}
\label{Chap performance bound}
Combining the theoretical D-R performance and the mapping polynomial, we can have a mapping relation between the bit rate and the LSD. As the D-R performance is the lower-bound of VQ, the VQ performance bound, in terms of the LSD, can be obtained analytically. The TIMIT database~\cite{TIMIT1990} was used to generate a set of $16$-dimensional LSF parameters. With~\eqref{dLSF}, the corresponding $\Delta$LSF parameter vectors were obtained. With window length equal to $25$ milliseconds and step size equal to $20$ milliseconds, approximate $706k$ training $\Delta$LSF vectors were extracted. The Hann window was applied to each frame and no prefilter was used. All the silent frames were removed.

\begin{figure}[!t]
     \psfrag{x}[][]{ \scriptsize Rate (bits/vector)}
    \psfrag{y}[][]{\scriptsize LSD (in dB)}
 \vspace{-2mm}
%    \begin{tabular}{@{}cc@{}}
%
%     \subfigure[\sps Log-likelihood comparison]{\includegraphics[width=.224\textwidth]{D-R.eps}}& \subfigure[\sps BIC comparison]{\includegraphics[width=.224\textwidth]{LSD-R.eps}}
%\end{tabular}
\centering
\includegraphics[width=.4\textwidth]{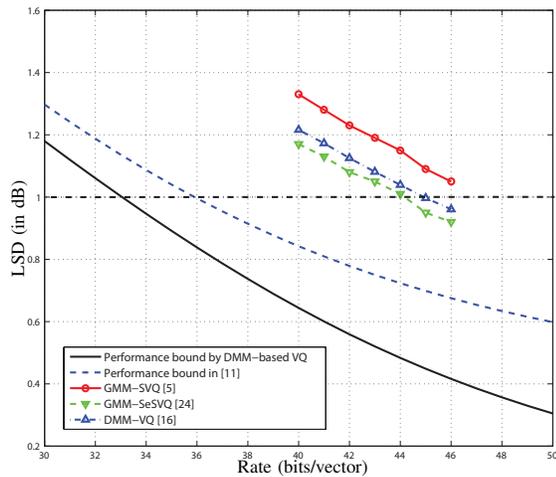}

\vspace{0mm}
          \caption{ \label{Fig:LSDR} \small Comparisons of the estimated performance bound with the bound estimated in~\cite{Chatterjee2008}. The LSD-rate cures obtained from different VQs are also plotted to illustrate the gaps between the practical performance and the bound.}
          \vspace{-3mm}
\end{figure}

We trained several DMMs with different model complexities (in terms of number of mixture components,~\emph{e.g.}, $64$, $128$, $256$) with the $\Delta$LSF vectors.  As shown in the figure, the difference between the curve with $128$ mixture components and the curve with $256$ components can be neglected. This indicates that increasing model complexity will not yield significant performance improvement. Hence, we applied $256$-component DMM-based VQ to get the D-R relation of the $\Delta$LSF parameters. With the transformation relation introduced in~\cite{Ma2010}, the distortion in the $\Delta$LSF domain was transformed to the LSF domain. Finally, the LSD-Rate performance of the DMM-based VQ was obtained, by the assistance of the third-order polynomial. The performance bound comparison between the DMM-based VQ and the GMM-based VQ is shown in Fig.~\ref{Fig:LSDR}. The minimum bit rate estimated by the DMM-based VQ is $33$ bits/vector. Compared to the performance bound estimated in~\cite{Chatterjee2008}, it is $3$ bits less for transparent coding. Moreover, there is about $10$ bits gap to be bridge between some previously implemented methods~\cite{Paliwal1993,Chatterjee2007,Ma2013} and the estimated bound. This indicate that there is large space to improve the practical VQ performance.
%\begin{figure}[!t]
%     \psfrag{x}[][]{ \scriptsize Rate (bits/vector)}
%    \psfrag{y}[][]{\scriptsize MSE}
%\vspace{-2mm}
%\centering
%
%\includegraphics[width=.4\textwidth]{D-R.eps}
%
%\vspace{0mm}
%          \caption{ \label{Fig:DR} \small D-R performance of DMM-based VQ with model complexity.}
%          \vspace{-3mm}
%\end{figure}

\section{Conclusion}
\label{Chap conclusion}
The performance bound of the LSF VQ is estimated via a mixture of Dirichlet distributions. According to the high rate assumption and with the constrained entropy case, the D-R performance of the DMM-based VQ is calculated with an analytically tractable form. By empirically mapping the MSE to the LSD, the relation between the LSD and the bit rate is obtained. The minimum bit rate required for transparent coding is illustrated. The gaps between the previously implemented VQs and the estimated bound indicates there exists large space for improving the performance of practical VQs.

\end{document}